\pdfoutput=1

\documentclass[11pt]{article}

\usepackage{ACL2023/ACL2023}

\usepackage{times}
\usepackage{latexsym}

\usepackage[T1]{fontenc}

\usepackage[utf8]{inputenc}

\usepackage{microtype}
\usepackage{inconsolata}

\usepackage{textcomp}
\usepackage[utf8]{inputenc}
\usepackage{amsmath,amssymb}
\usepackage[ruled]{algorithm2e}
\usepackage{algorithmic}
\usepackage{graphicx}
\usepackage{array}
\usepackage{booktabs}
\usepackage{float}
\usepackage{caption}
\usepackage{natbib}
\usepackage{cite}
\usepackage{multirow}
\usepackage{comment}
\usepackage{mathtools}

\usepackage{amsfonts}
\usepackage{url}

\setcitestyle{authoryear}

\usepackage{url,hyperref}
\usepackage{color}

\title{Enhancing Coherence of Extractive Summarization with Multitask Learning}

\author{Renlong Jie, Xiaojun Meng, Lifeng Shang, Xin Jiang, Qun Liu \\
  Huawei Noah’s Ark Lab \\
  \texttt{\{jierenlong, xiaojun.meng, Shang.Lifeng, Jiang.Xin, qun.liu\}@huawei.com} \\}

\begin{document}
\maketitle
\begin{abstract}
This study proposes a multitask learning architecture for extractive summarization with coherence boosting. The architecture contains an extractive summarizer and coherent discriminator module. The coherent discriminator is trained online on the sentence vectors of the augmented textual input, thus improving its general ability of judging whether the input sentences are coherent. Meanwhile, we maximize the coherent scores from the coherent discriminator by updating the parameters of the summarizer. To make the extractive sentences trainable in a differentiable manner, we introduce two strategies, including pre-trained converting model (\textit{model-based}) and converting matrix (\textit{MAT-based}) that merge sentence representations. Experiments show that our proposed method significantly improves the proportion of consecutive sentences in the extracted summaries based on their positions in the original article (\textit{i.e.,} automatic sentence-level coherence metric), while the goodness in terms of other automatic metrics (\textit{i.e.,} Rouge scores and BertScores) are preserved. Human evaluation also evidences the improvement of coherence and consistency of the extracted summaries given by our method.
\end{abstract}

\section{Introduction} \label{Sec:1}

Coherence is an important property to be concerned for text summarization, which is highly related to user experience and affects the difficulty in understanding the generated or extracted summaries. A lot of efforts have been paid in improving the coherence of summaries in existing studies \citep{Janara2013Towards, Daraksha2015Integrating, yuxiang2018learning, Nadeem2020Extractive}. 

Extractive summarization has a large potential in real-world application. Compared with abstractive summarization, it can better preserve the factual consistency with the original article, while the inference time is generally faster as autoregressive generation is not required \citep{liu2019fine, Wafaa2021Automatic}. Although some recent studies tend to apply GPT-series models for both abstractive and extractive summarization \citep{goyal2022news, zhang2023extractive}, the traditional encoder-only model has much smaller size, thus requiring less demand on computing resources. Therefore, encoder-based extractive summarization is often less costly when adopted in a real-time user application. 

Besides low level of abstractness, another urgent issue of using extractive summarization is that the coherence of the extracted sentences is generally not guaranteed \citep{Janara2013Towards, yuxiang2018learning}. When an article involves a number of sentences, the extractive model could select sentences long-way apart from each other. In addition, several consecutive sentences that have the causal, transition, progressive or juxtaposition relationships could be disjointed by extractive models, which not only harms the coherence, but also leads to factual inconsistency in extracted summaries.

To improve the coherence of extractive summarization models, there are two major strategies in existing studies. The first is to select many candidate summary sentences based on the score rankings, and then re-rank them by considering the coherent scores of these sentences \citep{Janara2013Towards, Daraksha2015Integrating}. The second is using reinforcement learning for improving a coherence reward~\citep{yuxiang2018learning}. 
However, few attentions are paid to use the state-of-the-art transformers for coherence enhancement in a end-to-end manner. 
In this paper, we propose a novel transformer-based multitask learning architecture for enhancing coherence in extractive summarization, which can be applied to any extractive summarization that score and select sentences, not only for sentence-level, but also phrase/sub-sentence level (separated with comma, etc.) extraction. Our main contributions are three-fold:
\begin{itemize}
\setlength{\itemsep}{0.5pt}
\setlength{\parsep}{0.5pt}
\setlength{\parskip}{0.5pt}
    \item We use a data augmentation technique by shuffling sentences from the input article.
    Trained on it, we introduce a novel transformer-based coherent discriminator that gives the coherent score of each sentence, and it can be also trained online with a general extractive summarizer to improve the coherence.
    \item We propose a joint multitask learning architecture with three objectives that can train the coherent discriminator and extractive summarizer simultaneously. 
    \item Extensive studies show that our proposed multitask method can significantly improve the coherence of an extracted summary. Human evaluation further evidences it, and also indicates that relevance and factual consistency can be promoted as a by-product.
\end{itemize}

\section{Related work} \label{Sec:2}

\subsection{Extractive Summarization} \label{Sec:2.1}
Extractive summarization is to extract a set of text units to formulate a short summary for a given long article. 
Traditional extractive summarization methods like Textrank and Lexrank \citep{mihalcea2004textrank, erkan2004lexrank} compute the importance of each sentence based on its centrality in the whole article. By using neural networks, extractive summarization can be considered as a sequence labelling problem, where the model scores each sentence for ranking and selecting \citep{neuralsumm2016}. In the era of transformer, pretrained language models are widely adopted to further improve the extractive summarization. One of the most popular ones is BertSum~\citep{liu2019fine}, which uses BERT~\citep{bert} as the text encoder, and fine-tunes the \texttt{[CLS]} token at the beginning of each sentence for scoring~\citep{liu2019fine}.  Graph-based neural networks~\citep{hetergraph2020, multiplexgraph2021} are also used to model the inter-sentence or sentence-word relationships, where different levels of granularity can be modelled as semantic nodes in a heterogeneous graph. \citet{nested2021} further considers nested tree structures of documents to explicitly represent sentence information. \citet{gu2022memsum} introduces a reinforcement-learning-based extractive summarizer enriched with information on the current extraction history at each step. 



\subsection{Coherence Model} \label{Sec:2.2}
Coherent models are used to distinguish a coherent text from an incoherent sequence of sentences. For example, Entity Grid represents an article by a two-dimensional array, which captures transitions of discourse entities across sentences \citep{regina2008modeling}. \citet{extent2011} further extends it by adding entity-specific features such as discourse prominence, named
entity type and coherence features. \citet{neuralcoherence2017} introduces a convolutional neural network over the entity grid representations to model local coherence. In addition, contextual word embeddings like ELMO are added to the entity-grid feature to further improve coherence models~\citep{Shafiq2018cohe, han2019a}. Generally, previous methods are trained on a global discrimination task, which aims to distinguish an original coherent document from its incoherent counterparts by random shuffling its sentences \citep{regina2008modeling}. However, this global discrimination task does not provide an effective coherence model for downstream tasks such as abstractive and extractive summarization \citep{mohiuddin2021rethinking, laban2021can}. 

\begin{figure*}[th]
\begin{center}
 \includegraphics[width=0.8\linewidth]{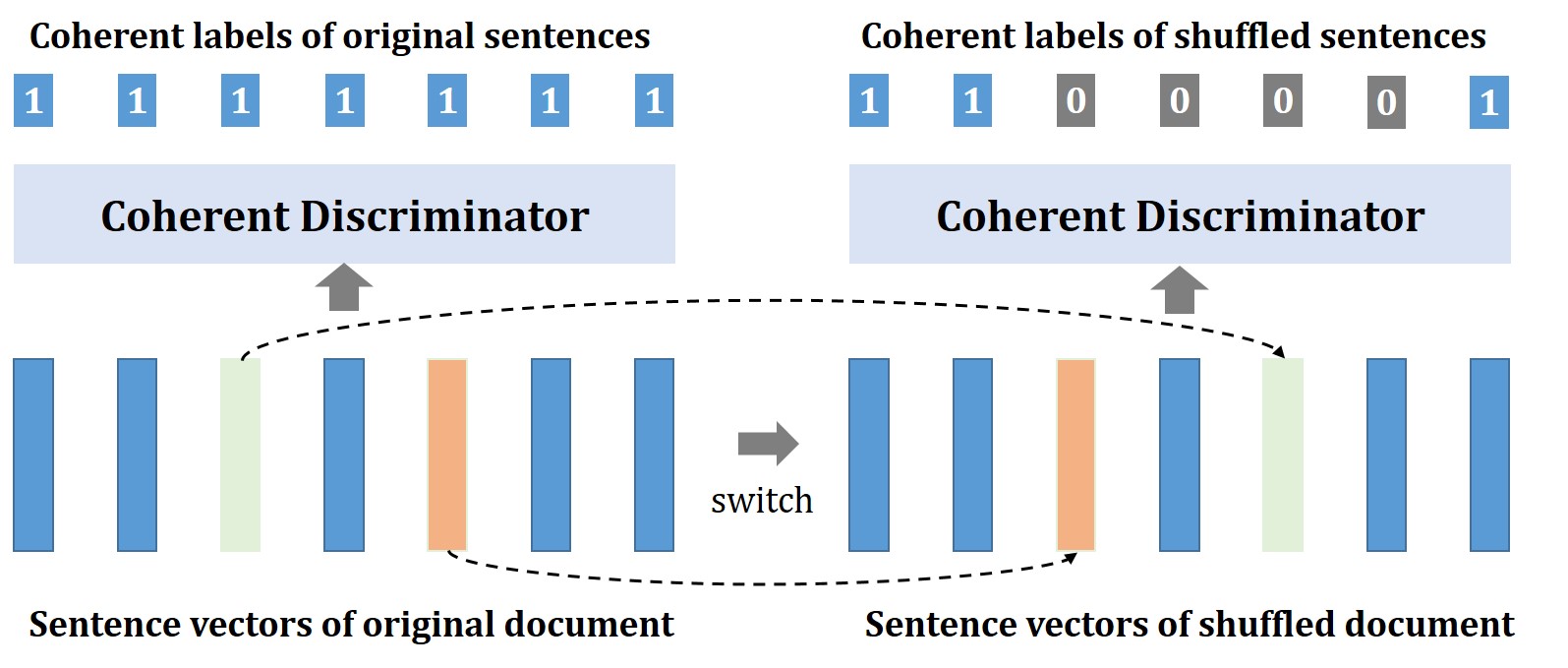}
 \caption{Diagram of transformer-based coherent discriminator on shuffled sentences. Its input is the sentence vector \texttt{[CLS]} from BERT. The coherent label of each sentence is 1 if it still follows its prior sentence from the original document after sentence shuffling. The coherent discriminator is trained online with summarization modules in the multitask manner.} 
 \label{Fig:coherence}
\end{center}
\vspace{-1em}
\end{figure*}


\subsection{Coherence Enhancing in Summarization} \label{Sec:3.2}
Existing approaches are studied on coherence enhancement for both abstractive and extractive summarization. In extractive method, G-FLOW~\citep{christensen_naacl13} introduces a joint model for selection and ordering that balances
coherence and salience by approximating the
discourse relations across sentences with a graph representation. \citet{Daraksha2015Integrating}
represent an input article by a bipartite graph consisting of sentence and entity nodes, and use a graph-based ranking algorithm to ensure non-redundancy and local coherence of the result summary. RNES \citep{yuxiang2018learning} applies reinforcement learning to capture the cross-sentence coherence patterns and optimizes the coherence and informative importance of the result summary simultaneously. However, this method relies on a pre-trained adjacency-based coherent model to provide the coherence reward for reinforcement learning \citep{hu2014convolutional}, which could involve low cross-domain/dataset adaptability and low training efficiency. Different from previous work, and as far as we know, we are the first to propose a transformer-based multitask architecture for end-to-end coherence enhancement in extractive summarizer.

\section{Method} \label{Sec:3}
In this paper, we propose a multitask learning model for supervised extractive summarization with coherent boosting. We introduce the main components of the proposed method as below.

\subsection{BertSum for Extractive Summarization}
As extractive summarization can be regarded as a sentence-level sequence labelling task, fine-tuning BERT by adding an extra sentence encoder has worked fairly well~\citep{liu2019fine}, given that BERT owns the strong context-aware capability of representing words and sentences.
Therefore, we also use BERT as the representation learner with all sentences as input, and use the vector of \texttt{[CLS]} token to represent each sentence. We add an extra transformer layer on top of the \texttt{[CLS]} vector to score each sentence, as similar to BertSum~\citep{liu2019fine}.

\subsection{Sentence Shuffling for data augmentation} \label{Sec:3.1}

We apply sentence shuffling to create incoherent examples. For each document, we sample the number of sentence switch $n$ from a Poisson distribution $\text{Possion}(\lambda)$, where $\lambda$ is a hyper-parameter showing the average number of switching, and $\lambda$ is shared across all documents. We then label the sentence not following its prior sentence from the original document as ``incoherent'', while the sentence still following its prior one as ``coherent''.

We introduce a coherent discriminator that aims to determine if an input textual sequence is coherent in a sentence level. It consists of transformer layers that take the sentence representation vector (i.e., \texttt{[CLS]}) from BERT as input, and output a score to measure the coherence of this sentence with its prior one. The overview method is presented in Figure~\ref{Fig:coherence}. 
There are two advantages of our proposed coherent discriminator. Firstly, as the model is trained in a supervised manner on every sentence, bringing it a larger possibility in well modelling coherence compared to previous discussed methods \citep{mohiuddin2021rethinking, laban2021can}. Secondly, instead of offline pre-training methods, we train the coherent discriminator online with summarization models as multi-tasking learning. 

\subsection{The joint Multi-task model} \label{Sec:3.3}
\begin{figure*}[th]
\begin{center}
 \includegraphics[width=0.95\linewidth]{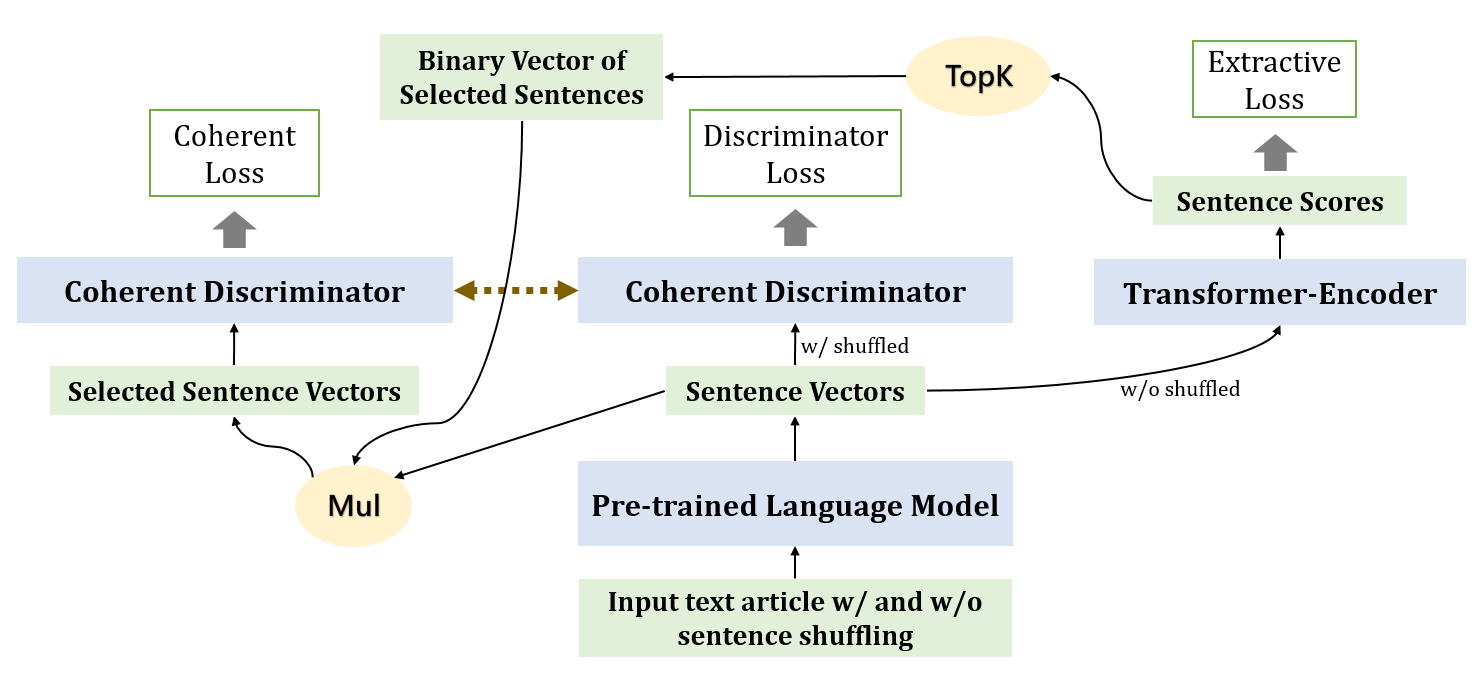}
 \caption{Architecture for proposed multitask coherent enhancement model for fine-tuning pre-trained BertSum. Dashed line means the weight sharing strategy. We use TopK sentence selection for an example. The gradients from Discriminator Loss only updates the parameters in the coherent discriminator. The gradients from Coherent Loss only pass to the BERT module through sentence scores.} \label{Fig:joint_mt}
\end{center}
\vspace{-1em}
\end{figure*}
The architecture of the multitask model is given in Figure~\ref{Fig:joint_mt}. We first feed the text article into BERT to output the sentence vectors \texttt{[CLS]}.
Secondly, these sentence representations are feed forward through three paths: 1) they pass through a transformer encoder and output importance scores for each sentence as in BertSum~\citep{liu2019fine}. 2) they pass through the coherent discriminator and output the coherent scores for each sentence. 3) they multiply with a binary vector meaning selecting or not selecting each sentence, which is given by using the below Gumble-Softmax and TopK on importance scores from 1) \citep{jang2017categorical}. 
\begin{equation*}
\setlength{\abovedisplayskip}{3pt}
\setlength{\belowdisplayskip}{3pt}
    \begin{split}
        y_i &= \text{GumbleSoftmax}(\pi_i, T)\\
        &=\text{Softmax}(g_i + log \pi_i, T),\\
        i &=1,2,...,M\\
        s_{tk} & = \text{TopK}(\mathbf{y})
        \end{split} 
\end{equation*}
where $s_{tk}$ is a binary vector that means to select or not select that sentence as the final summary result, $\pi_i$ is the normalized sentence score, and $T$ is the temperature of softmax function. $g_i$ follows a gumble distribution
\begin{equation*}
\setlength{\abovedisplayskip}{3pt}
\setlength{\belowdisplayskip}{3pt}
  g_i = -\log(-\log(u_i)),\, u_i \sim U(0,1)
\end{equation*}

Indeed, there are alternatives to the above solution, such as directly using TopK or using 
the Gumble TopK sampling with:
\begin{equation*}
\setlength{\abovedisplayskip}{3pt}
\setlength{\belowdisplayskip}{3pt}
y_i = g_i + \log \pi_i,\, i =1,2,...,M
\end{equation*}
In our experiment, we make a comparison between TopK, Gumble TopK and Gumble-Softmax TopK for sampling the binary vector for sentence indices $y$ in training. After obtaining $y$, the following straight-through trick can be implemented:
\begin{equation*}
\setlength{\abovedisplayskip}{3pt}
\setlength{\belowdisplayskip}{3pt}
    \hat{\mathbf{s}} = (\mathbf{s}_{tk} - \mathbf{y})_f + \mathbf{y}
\end{equation*}
where $(.)_f$ means 
that 
the terms in the parentheses are fixed without gradients in back-propagation. Thus we obtain $\hat{s}$ as a differentiable binary vector, which can be applied to multiply with sentence vectors $e$ for the selecting process.
\begin{equation}
    \mathbf{e}'_i = \hat{s}_i \mathbf{e}_i,\quad s_i \in \{0,1\},\, i\in \{1,...,n\}
\label{Eq:1}
\end{equation}
where $\hat{s}_i$, $e_i$ and $\mathbf{e}'_i$ are the $i$-th element in $\hat{s}$, the representation vector for $i$-th sentence, and the new representation vector for $i$-th sentence after multiplication. Since the extractive summarization involves selecting sentences to form the final summary,
we use a novel representation merging operation to ensure the differentiability of this process,
which will be further discussed in Section~\ref{Sec:3.4}.

\paragraph{Training Objectives.} As shown in Figure~\ref{Fig:joint_mt}, we plot two of coherent discriminators. The one with \textit{Discriminator Loss} is trained on shuffled sentences from the original article to obtain the ability of judging whether its input sentences are coherent. The other one with \textit{Coherent Loss} is to calculate and minimize the negative coherent score of selected sentences without updating the discriminator, where the coherent score is the average of the output scores of the coherent discriminator on all its input sentence vectors. These two discriminators share the same weights but work for different sub-tasks in training. As the training process enhances the coherence by updating the extractor, the sentence representations change over time in training. Thus, the discriminator loss should be trained along with the coherent loss for online adaptation. 

Overall, in our multitask learning, there are three optimization objects: (a) Cross entropy loss of the scoring of extractive summarizer, i.e., \textit{Extractive Loss}, which is denoted as $L_{ext}$; (b) Cross-entropy loss of the coherent discriminator (between its output scores and coherent labels) for training this discriminator, i.e., \textit{Discriminator Loss}, which is denoted as $L_{dis}$; (c) Minimizing the negative coherent scores of selected sentences (with fixed discriminator) for coherence enhancing, i.e., \textit{Coherent Loss}, which is denoted by $L_{cohe}$. The total loss is as follows, where $\lambda_{ext}$, $\lambda_{dis}$ and $\lambda_{cohe}$ are hyper-parameters to be tuned.
\begin{equation}
\begin{split}
L = \lambda_{ext} L_{ext} + \lambda_{dis} L_{dis} + \lambda_{cohe} L_{cohe}
\end{split}
\end{equation}

\paragraph{Distinguish the forward and backward paths.}
Note that not all the parameters of shared coherent discriminators are updated with each loss.
We plot two coherent discriminators in the Figure~\ref{Fig:joint_mt}, to clearly distinguish two different forward and backward paths, where the actual implementation can be simpler.
In particular, the coherent discriminator with $L_{dis}$ is trained on labelled and shuffled sentence data of the entire article, while its weights are frozen in back-propagating the $L_{cohe}$, where the Discriminator Loss of only selected sentences are computed. It means that $L_{cohe}$ does not affect parameters of any coherent discriminator.
In addition, we also make $L_{cohe}$ update the parameters in pre-trained model, i.e., BERT-based extractor, only through sentence scores $\hat{s}$ instead of sentence vectors $\mathbf{e}_i$ in Eq.~\eqref{Eq:1}. This design is to avoid the shortcut of directly affecting the embedding of sentence vectors in BERT, which potentially harms and limits the capability of our coherent discriminator.
Table~\ref{tab:1} shows how gradients from different losses update parameters of different modules.

\begin{table}[htbp]
  \centering{\footnotesize{
    \begin{tabular}{cccc}
    \toprule
          & $L_{ext}$ & $L_{dis}$ & $L_{cohe}$ \\
    \midrule
    Extractor & T     & F     & T \\
    Cohe. Discriminator & F     & T     & F \\
    \bottomrule
    \end{tabular}%
      \caption{Parameter Updating by Different Losses. T denotes to TRUE and F denotes to FALSE.}
  \label{tab:1}}}%
  \vspace{-1em}
\end{table}%

\subsection{Representation merging for selected sentences} \label{Sec:3.4}
To obtain selected sentence representations in our multitask learning, it requires a differentiable selecting process on sentences. To address it, we explore in two settings. In the first \textit{MAT-based} setting, we directly use a transformation matrix $\mathbf{M} \in \mathcal{R}^{n\times m}$ to align the selected sentence representation $\mathbf{E}^{'}\in \mathcal{R}^{n\times d}$ to a dense representation $\mathbf{E}_d \in R^{m\times d}$, where $n$ is the number of sentences in the original input text, $m$ is the number of selected sentences: 
\begin{equation}
\setlength{\abovedisplayskip}{4pt}
\setlength{\belowdisplayskip}{4pt}
\begin{split}
\mathbf{E}_d^{T} &= (\mathbf{e}_1, \mathbf{e}_2, \mathbf{e}_5) = \mathbf{E}^{'}{}^{T} \mathbf{M} \\ &= (\mathbf{e}_1, \mathbf{e}_2, \mathbf{0}, \mathbf{0}, \mathbf{e}_5)\begin{pmatrix}
1 & 0 & 0\\
0 & 1 & 0 \\
0 & 0 & 0 \\
0 & 0 & 0 \\
0 & 0 & 1
\end{pmatrix}
\end{split}
\label{Eq:3}
\end{equation}
As $\mathbf{E}^{'}$ fully determines the selected dense representation $\mathbf{E}_d$, $\mathbf{M}$ is not necessary to be differentiable in the gradient graph. The obtained $\mathbf{E}_d$ is further passed to the coherent discriminator to calculate the coherent scores of selected sentences. 

In the second \textit{model-based} setting, we apply a pre-trained transformer encoder to achieve selected representation merging, by mapping the binary vector $\hat{\mathbf{s}} \in \mathcal{R}^{1\times n}$ to a conversion matrix $\mathbf{M}\in \mathcal{R}^{n\times m}$, where $\hat{\mathbf{s}}$ is the binary vector after top K selection, and $\mathbf{M}$ is the same converting matrix in Eq.~\eqref{Eq:3}. 
\begin{equation}
\setlength{\abovedisplayskip}{3pt}
\setlength{\belowdisplayskip}{3pt}
    \mathbf{M} = f(\hat{\mathbf{s}})
\end{equation}
This transformer encoder includes 3 layers with 6 attention heads and hidden size of 30. We first use a linear transformation $\mathbf{W}_{in} \in \mathcal{R}^{1\times d}$ to convert the binary vector $\hat{\mathbf{s}}$ to the input dimension of this transformer encoder, and use another linear transformation $\mathbf{W}_{out} \in \mathcal{R}^{d\times M}$ on the output to form a matrix with size $n\times m$. To pre-train such a transformer, we sample a training set of binary vectors, and their lengths (i.e., the simulated number of sentences) follow a Poisson distribution $\text{Poisson}(\lambda)$, where $\lambda$ is the average number of sentences within the first 512 tokens of the summarization dataset. We randomly select indices of the sampled binary vectors and set them as 1, while the number of selected indices equal to the target number of extractive sentences (3 by default). 

\section{Experiments} \label{Sec:4}


\subsection{Dataset} \label{Sec:4.1}
We evaluate our proposed method three widely-used datasets: CNNDM~\citep{hermann2015teaching}, NYT~\citep{durrett2016learning} and CNewSum~\citep{wang2021cnewsum}.
CNNDM~\citep{hermann2015teaching} contains news articles from the CNN and Daily Mail websites, with labelled abstractive and extractive summaries. There are 287,226 training samples, 13,368 validation samples and 11,490 test samples. NYT~~\citep{durrett2016learning} contains 110,540 article with abstractive summary. We follow its paper to split the original dataset into 100,834 training and 9,706 test examples. The extractive labels of sentence indices are generated by ORACLE approach~\citep{durrett2016learning}. CNewSum~ \citep{wang2021cnewsum} a Chinese news article dataset with labelled abstractive and extractive summaries. It contains 275,596 training samples, 14,356 validation samples and 14,355 test samples.

\subsection{Evaluation Criterion} \label{Sec:4.2}

In our experiment, we apply both automatic and human evaluation. For automatic evaluation, Rouge scores \citep{rouge2004package} and BertScore (F1-scores) \citep{zhang2019bertscore} are the criterion for comparing the performance of candidate models. 
These two automatic metrics evaluate the goodness of summaries by syntactical and semantic similarity, respectively. To evaluate the coherence of extracted summaries, we propose to apply the proportion of consecutive sentences in a result summary based on their positions in the original article. For example, consider an extracted summary with $N$ sentences, assuming the starting point and the first sentence is also ``consecutive'', then there are $N$ pairs of consecutive sentences. If $M$ pairs of them are also consecutive in the original article, the proportion of consecutive sentences is $M/N$.
Previous work on measuring the coherence of summaries mainly relies on human evaluation~\citep{yuxiang2018learning}, which is also applied in our study. 

\subsection{Implementation} \label{Sec:4.3}


We have discussed some of the implementation details in Section~\ref{Sec:3}, including the forward and backward paths of the three losses. For the coherent discriminator, we use a transformer encoder with 6 layers, 8 attention heads and hidden size of 768. 
Details of the converting matrix for merging sentence representations are discussed in Section~\ref{Sec:3.4}.
We apply Adam optimizers with $\beta_1 = 0.9$, $\beta_2 = 0.999$, and the learning rate is 2e-5 without schedule for all sub-tasks. To avoid involving too many factors, we do not apply Block Trigram trick~\citep{liu2019fine} for extractive summarization. 
We apply a two-stage training procedure: In the first stage, we set $L = L_{ext}$ to pretrain our Bertsum module on the original unshuffled training dataset. In the second stage, we apply the multitask fine-tuning on shuffled training dataset and set $\lambda_{dis}=\lambda_{cohe}=1.0$. Preliminary experiments show that in the fine-tuning stage, our framework can work fairly well even without using $L_{ext}$. To make the training process easy and efficient, we can simply set $\lambda_{ext}=0$. For the validation and test, we use the original unshuffled validation dataset for summary extraction.

\subsection{Results} \label{Sec:4.4}

\subsubsection{Automatic evaluation} \label{Sec:4.4.1}
\label{automatic_evaluation}
For each setting, we select the checkpoint with the maximum value of the summation over a set of evaluation metrics including Rouge1, Rouge2, RougeL, BertScore and the proportion of consecutive indices (i.e., \textit{Cons. Prop.}) in the original text, which are all ranged between 0 and 100 in percentage. We denote this summation metric as ``\textit{S-score}'', and evaluate the performance of selected checkpoints on the test set and report the median of three trials. We design this criterion to balance both automatic summarization and coherence metrics to select a checkpoint on the validation set. We use a summation form to involve each aspect while avoiding extra hyper-parameter tuning. 
In addition, we can use a weighted linear combination of all these metrics, thus highlighting a particularly concerned one, for which some extra results are given in Appendix~\ref{sec:appendix 1}. 
In the tables, R1, R2, RL, B.S., Cons.Prop. refer to Rouge-1, Rouge-2, Rouge-L, BertScore, and Consecutive Proportion, respectively.

\paragraph{Baseline.}
As discussed in Section~\ref{Sec:3.4}, there are two ways of merging the representations of selected sentences to make it differentiable. We believe it could make a difference on the model performance. We perform multitask fine-tuning with pre-trained BertSum modules on three datasets, and compare the BertSum baseline with our methods using two settings of representation merging (e.g., \textit{MAT-based} \& \textit{Model-based}).

\begin{table}[htbp]
  \centering{\footnotesize
    \begin{tabular}{p{3.5em}p{2.0em}p{1.3em}p{1.3em}p{1.3em}p{1.3em}p{1.3em}p{1.8em}}
    \toprule
          \textbf{Dataset} & \textbf{Setting} & \textbf{R1} & \textbf{R2} & \textbf{RL} &  \textbf{B.S.}  & \textbf{Cons. Prop.} & \textbf{S-score} \\
    \midrule
    \multirow{3}[2]{*}{CNewSum} & Base  & 34.6  & 18.5  & 29.1  & 71.9  & 61.6  & 215.5  \\
    & MAT   & 34.6  & 18.5  & 29.1  & 71.9  & 60.8  & 214.7  \\
    & Model & 34.5  & 18.5  & 29.1  & 71.8  & \textbf{66.2} & \textbf{220.0 } \\
    \midrule
    \multirow{3}[2]{*}{CNNDM} & Base  & 42.1  & 19.3  & 38.6  & 63.6  & 46.1  & 209.4  \\
    & MAT   & 41.5  & 18.4  & 38.1  & 62.9  & \textbf{72.8 } & \textbf{233.6 } \\
    & Model & 42.0  & 19.3  & 38.6  & 63.5  & 45.6  & 208.5  \\
    \midrule
    \multirow{3}[2]{*}{NYT} & Base  & 45.1  & 25.1  & 40.7  & 63.5  & 57.8  & 231.9  \\
    & MAT   & 44.8  & 24.8  & 40.4  & 63.4  & \textbf{65.3}  & \textbf{238.7} \\
    & Model & 45.0  & 25.0  & 40.6  & 63.3  & 58.0  & 231.7  \\
    \bottomrule
    \end{tabular}%
      \caption{Comparison of the baseline Bertsum and our multitask methods using two merging settings. }
  \label{tab:merge}}%
\end{table}%

\paragraph{Result.} As shown in Table~\ref{tab:merge}, we find that for all datasets, the proportion of consecutive sentences can be significantly improved using our method. Meanwhile, the Rouge scores and BertScore are generally not affected in a large scale. Note that the two merging settings are not effective for all cases. We observe that for Chinese CNewSum dataset, the \textit{model-based} setting performs better, while for English ones (CNNDM and NYT), the \textit{MAT-based} setting is better. We believe the selection of two settings can be considered as a hyper-parameter, which depends on the nature of datasets (\textit{e.g.,} language, sentence length). Based on this finding, we use \textit{model-based} for the Chinese dataset CNewSum and \textit{MAT-based} for the other two English datasets in the following experiments. Overall, our proposed multitask method is able to find a trade-off between preserving the traditional automatic metrics on the quality of a summary, while still improving its proportion of consecutive sentences. An further graphical analysis of learning curves and positional distribution of selected sentences are given in Appendix~\ref{Sec:graph_analysis}.



\begin{table}[htbp]
  \centering{\footnotesize
    \begin{tabular}{p{3.5em}p{2.0em}p{2.0em}p{1.3em}p{1.3em}p{1.3em}p{1.8em}}
    \toprule
    \textbf{Dataset} & \textbf{Num. Sent.} & \textbf{Setting} & \textbf{RL} & \textbf{B.S.} & \textbf{Cons. Prop.} & \textbf{S-score} \\
    \midrule
    \multirow{6}[1]{*}{CNewSum} & \multirow{2}[1]{*}{K=3} & Base  & 29.1  & 71.9  & 61.6  & 215.5  \\
          &       & Model & 29.1  & 71.8  & \textbf{66.2 } & \textbf{219.9 } \\
          & \multirow{2}[0]{*}{K=4} & Base  & 27.0  & 71.5  & 68.4  & 215.6  \\
          &       & Model & 26.9  & 71.3  & \textbf{72.8 } & \textbf{219.3 } \\
          & \multirow{2}[0]{*}{K=5} & Base  & 25.5  & 71.1  & 73.5  & 215.8  \\
          &       & Model  & 25.5  & 71.1  & 73.5  & 215.8  \\
    \midrule
    \multirow{6}[0]{*}{CNNDM} & \multirow{2}[0]{*}{K=3} & Base  & 38.6  & 63.6  & 46.1  & 209.4  \\
          &       & MAT  & 38.1  & 62.9  & \textbf{72.8 } & \textbf{233.6} \\
          & \multirow{2}[0]{*}{K=4} & Base  & 39.1  & 63.9  & 54.0  & 218.7  \\
          &       & MAT  & 39.0  & 63.5  & \textbf{71.5 } & \textbf{235.2 } \\
          & \multirow{2}[0]{*}{K=5} & Base & 38.7  & 63.7  & 59.1  & 222.2  \\
          &       & MAT  & 38.6  & 63.4  & \textbf{75.1 } & \textbf{237.3 } \\
    \midrule
    \multirow{6}[1]{*}{NYT} & \multirow{2}[0]{*}{K=3} & Base & 40.7  & 63.5  & 57.8  & 231.9  \\
          &       & MAT & 40.4  & 63.4  & \textbf{65.3} & \textbf{238.7} \\
          & \multirow{2}[0]{*}{K=4} & Base  & 40.7  & 63.3  & 59.4  & 232.5  \\
          &       & MAT  & 39.6  & 62.6  & \textbf{76.9} & \textbf{246.0} \\
          & \multirow{2}[1]{*}{K=5} & Base  & 40.1  & 62.9  & 63.4  & 233.7  \\
          &       & MAT & 37.9  & 61.6  & \textbf{81.5 } & \textbf{244.3 } \\
    \bottomrule
    \end{tabular}%
      \caption{Comparison of the baseline Bertsum and our multitask methods when varying the target number of extractive sentences.}
  \label{tab:n_sent}}%
\end{table}%

To check the performance of our proposed model when working with different number of selected sentences, we perform another experiment varying the sentence number $K=3,4,5$. As shown in Table~\ref{tab:n_sent}, we learn that for all datasets, \textit{S-score} and the \textit{Cons. Prop.} using our proposed method can be significantly increased with a large improvement for various sentence numbers except CNewSum when $K=5$, where we observe the validation \textit{S-score} decreases as the training moves on. Overall, the above results demonstrate the effectiveness of our proposed multitask method.

\subsubsection{Ablation Study on Training Workflow.} \label{Sec:4.4.3}

We design the training workflow of our method given in Table~\ref{tab:1} based on intuition and model analysis. 
We further check the effect of each setting with an ablation study. The results are given in Table~\ref{tab:A2}, in which ``base'' refers to the baseline pre-trained models without coherence enhancement, and ``default'' refers to the default setting of coherence enhancing model as in Section~\ref{Sec:4.4.1}. We consider the case that the gradients from $L_{dis}$ update the extractor in multitask fine-tuning, the case that $L_{cohe}$ updates the discriminator as well as the extractor, and the case that $L_{cohe}$ updates the extractor through sentence vectors as well as sentence scores. It is shown that on both CNNDM and NYT, other settings apart from our default workflow designing will result in a reduction of \textit{S-score}. 

\begin{table*}[htbp]
  \centering{\footnotesize
    \begin{tabular}{p{12em}|p{1.3em}p{1.3em}p{1.3em}p{1.3em}p{1.3em}p{1.5em}|p{1.3em}p{1.3em}p{1.3em}p{1.3em}p{1.3em}p{1.5em}}
    \toprule
    \multirow{2}[4]{*}{\textbf{Ratio}} & \multicolumn{6}{c|}{\textbf{CNNDM}}                  & \multicolumn{6}{c}{\textbf{NYT}} \\
\cmidrule{2-13}          & \textbf{R1}    & \textbf{R2}    & \textbf{RL}    & \textbf{B.S.}  & \textbf{Prop. Conti.} & \multicolumn{1}{c|}{\textbf{S-score}} & \textbf{R1}    & \textbf{R2}    & \textbf{RL}    & \textbf{B.S.}  & \textbf{Prop. Conti.} & \multicolumn{1}{c}{\textbf{S-score}} \\
    \midrule
    Base  & 42.0  & 19.3  & 38.6  & 63.5  & 46.0  & 209.4 & 45.1  & 25.1  & 40.7  & 63.5  & 57.8  & 231.9 \\
    Default & 41.5  & 18.4  & 38.1  & 62.9  & 72.8  & \textbf{233.6 } & 44.8  & 24.8  & 40.4  & 63.4  & 65.3  & \textbf{238.7} \\
    $L_{dis}$ updates extractor & 41.4  & 18.8  & 38.1  & 62.9  & 58.6  & 219.7  & 44.0  & 24.1  & 39.6  & 62.2  & 68.0  & 237.9  \\
    $L_{cohe}$ updates discriminator & 41.4  & 18.5  & 38.1  & 62.9  & 71.8  & 232.7  & 44.8  & 24.8  & 40.5  & 63.2  & 63.7  & 237.1  \\
    $L_{cohe}$ updates sent. vecs. & 40.8  & 18.0  & 37.5  & 62.3  & 67.9  & 226.5  & 43.6  & 23.7  & 39.1  & 62.4  & 66.7  & 234.1  \\
    \bottomrule
    \end{tabular}%
    \caption{Ablation study on the workflow of multitask model training.}
  \label{tab:A2}}
\end{table*}%

\subsubsection{Selection of sampling methods} \label{Sec:4.4.2}

In Section~\ref{Sec:3.4}, we discuss the potential sampling methods of selected sentence to create simulated training data. The basic method is topK method, which is usually applied in inference phase for extracting sentences. However, in training phase, adding a random sampling process with a certain level of noises could potentially benefit the training. Thus, we perform an experiment to compare the cases of using three different sampling methods. As shown in Table~\ref{tab:sample_method}, we can see that in general, all the three methods can result in an improvement in terms of \textit{S-score}, while Gumble topK (\textit{\textbf{GK}}) and Gumble-Softmax topK (\textit{\textbf{GSK}}) perform better than \textit{\textbf{topK}} method. Gumble-Softmax topK gets the highest \textit{S-score} on CNewSum, while Gumble topK do better on CNNDM and NYT data-sets but with a significant decrease of relevance scores. Therefore, the reported results in our main experiments (Section~\ref{automatic_evaluation} and~\ref{human_evaluation}) adopt the Gumble-Softmax topK as our sampling method.
\begin{table}[htbp]
  \centering{\footnotesize
    \begin{tabular}{p{3.5em}p{2.0em}p{1.3em}p{1.3em}p{1.3em}p{1.3em}p{1.3em}p{1.8em}}
    \toprule
    \textbf{Dataset} & \textbf{Method} & \textbf{R1} & \textbf{R2} & \textbf{RL } & \textbf{B.S.} & \textbf{Cons. Prop} & \textbf{S-score} \\
    \midrule
    \multirow{4}[2]{*}{CNewSum} & Base  & 34.6  & 18.5  & 29.1  & 71.9  & 61.6  & 215.6  \\
          & TopK    & 34.4  & 18.4  & 29.0  & 71.7  & 64.6  & 218.0  \\
          & GK    & 34.5  & 18.5  & 29.1  & 71.8  & 64.8  & 218.5  \\
          & GSK    & 34.5  & 18.5  & 29.1  & 71.8  & 66.2  & \textbf{219.9 } \\
    \midrule
    \multirow{4}[2]{*}{CNNDM} & Base  & 42.1  & 19.3  & 38.6  & 63.6  & 46.1  & 209.5  \\
          & TopK    & 42.2  & 19.4  & 38.7  & 63.6  & 47.4  & 211.1  \\
          & GK    & 40.9  & 18.0  & 37.6  & 62.4  & 82.4  & \textbf{241.2 } \\
          & GSK    & 41.5  & 18.4  & 38.1  & 62.9  & 72.8  & 233.7  \\
    \midrule
    \multirow{4}[2]{*}{NYT} & Base  & 45.1  & 25.1  & 40.7  & 63.5  & 57.8  & 231.9  \\
          & TopK    & 42.4  & 22.8  & 37.9  & 61.8  & 73.6  & 238.3  \\
          & GK    & 43.2  & 23.3  & 38.7  & 62.3  & 74.4  & \textbf{241.7}  \\
          & GSK    & 44.8  & 24.8  & 40.4  & 63.4  & 65.3  & 238.7 \\
    \bottomrule
    \end{tabular}%
     \caption{Comparison of the baseline Bertsum and our models using three types of sampling methods. GK denotes to Gumble topK and GSK denotes to Gumble-Softman topK.}
  \label{tab:sample_method}}
  \vspace{-1em}
\end{table}%

\subsubsection{Human evaluation} 
\label{human_evaluation}
We further conduct a manual evaluation on the CNNDM and NYT dataset to check (a) if our proposed method improves the coherence of extracted summaries and (b) if it potentially reduces the other aspects of summaries such as relevance and factual consistency. We randomly sample 1000 entries from the test set and generate summaries with the baseline BertSum and our proposed multitask method.
Each sample is scored between 1 and 5 points by three participants from Amazon Mechanical Turk to answer three questions: (a) whether the summary captures the key points of the source document (Relevance); (b) the sentence-level coherence of the summary (Coherence); (c) the factual consistency of the summary (Consistency). 

Results of average ratings are reported in Table~\ref{tab:human}. For all settings, the standard error of the sample mean is around 0.015 to 0.017. Based on $2\sigma$ rule, it apparently indicates that our method provides a significant improvement of most metrics over the strong baseline Bertsum, not only in terms of the coherence but also 
the relevance and consistency.
Therefore, results evidence that our proposed method can bring an improvement of coherence without reducing other aspects of the summary quality. We believe it also brings other benefits, like the improvement of consistency, which may be due to that our method reduces the possibility of co-reference mismatch that might often happen when extracting in-consecutive sentences.

\begin{table}[htbp]
  \centering{\footnotesize
    \begin{tabular}{ccccccc}
    \toprule
          & \multicolumn{3}{c}{\textbf{CNNDM}} & \multicolumn{3}{c}{\textbf{NYT}} \\
\cmidrule{2-7}          & \textbf{Rel.} & \textbf{Coh.} & \textbf{Con.} & \textbf{Rel.} & \textbf{Coh.} & \textbf{Con.} \\
\cmidrule{2-7}    Base  & 3.34  & \textbf{3.29}  & 3.32  & 3.22  & 3.26  & 3.28 \\
    Ours  & \textbf{3.39}  & 3.28  & \textbf{3.38}  & \textbf{3.24}  & \textbf{3.31}  & \textbf{3.32} \\
    \bottomrule
    \end{tabular}%
     \caption{Human evaluation of different model outputs. Rel., Coh., Con. refer to relevance, coherence and factual consistency, respectively. We provide the average scores judged by human annotators. The improvements of consistency are significant for both datasets. No significant difference of coherence observed for CNNDM.}
  \label{tab:human}}%
  \vspace{-1.0em}
\end{table}%







\section{Discussion} \label{Sec:5}

In Section~\ref{Sec:4}, we apply the proportion of consecutive sentences in the original article as the automatic metric. Although non-consecutive sentences can also be locally coherent, the coherence of consecutive ones can be guaranteed given the original article is coherent. This could also contribute to the boosting of factual consistency as discussed above. 
Another advantage of our method is that we can train the model and select a checkpoint based on the emphasis on which of coherence and traditional evaluation metrics.
In fact, the weighting and combination of our used 5 metrics (\textit{i.e.,} R1, R2, RL, B.S., Cons.Prop) for training and validating can be further tuned according to real needs in practice as is discussed in Appendix~\ref{sec:appendix 1}. 

\paragraph{Comparison to RNES.}
The most similar coherence enhancing model to us is RNES \citep{yuxiang2018learning} that designs a coherence reward for reinforcement learning, and it only provides result on CNNDM ($R1=40.95$; $RL=37.41$). Compared to which, our method can achieve faster training for transformer-based models, generally less performance drop on automatic metrics like R1 \& RL, and factual consistency improvement in human evaluation. 

\section{Conclusion} \label{Sec:6}

This study provides an approach to enhance the coherence of extractive summarization with a multitask learning. We apply sentence permutation to generate labelled incoherent examples for training a coherent discriminator, which is also used to help the general summarization module on selecting coherent summary sentences. Two techniques for making the process of selecting sentences differentiable are introduced and compared. Automatic evaluation shows that our proposed method can significantly improve the proportion of consecutive sentences according to the positions in the original article, while the traditional automatic metrics of summary performance are preserved. Further, human evaluation evidences that the scoring performance of relevance and consistency can also be improved by adopting our method. 



\bibliography{Reference}

\newpage

\appendix

\section{Different ratios of consecutive proportion for checkpoint selection} \label{sec:appendix 1}

In Section~\ref{Sec:4.4}, we introduced a ``\textit{S-score}'' for checkpoint selection, where a summation form of 5 automatic metrics are summed up to avoid extra hyper-parameter searching. In practice, we may be in favor of either the traditional relevance metrics (i.e. Rouge scores and BertScore) or the coherence metric. Therefore, we also explore a more general form that is:
\begin{equation*}
\begin{split}
\text{S-score} &= \text{Rouge-1} + \text{Rouge-2} + \text{Rouge-L}\\ &+ \text{Bertscore}
+ \alpha\cdot\text{Cons. Prop.}
\end{split}
\end{equation*}
where $\alpha$ is a hyper-parameter to control the ratio of the consecutive proportion. To place emphasis on the relevance, we can apply a smaller value for $\alpha$, while in some scenarios that emphasize more on the coherence, we can apply a large value for $\alpha$ instead. 

By changing the value of $\alpha$ for selecting different checkpoints to be evaluated, we can get the results shown in Table ~\ref{tab:A1}. Note that this checkpoint selection for a particular dataset is relying on the same validation curve. Therefore, when using different ratios of $\alpha$, if the same optimal \textit{S-score} are obtained in the validation stage, then the same checkpoint as well as the same evaluation result are reported in Table ~\ref{tab:A1}.

We find that for CNNDM and NYT, when $\alpha$ is small, the performance on relevance metrics will be closer to the baseline model, while the consecutive proportion reduces but still significantly larger than that of baseline without coherence enhancement. For CNewSum, the relevance metrics will almost not reduce within the range of hyper-parameter settings of $\alpha$, but the consecutive proportion increases from $61$ to $66$ suddenly as $\alpha$ changes from $0.05$ to $0.1$. In general, it demonstrates that with the proposed coherence enhancing method, even a small sacrifice of relevance metric can still significantly improve the coherence of extracted summary.
\begin{table}[htbp]
  \centering{\footnotesize
    \begin{tabular}{p{3.5em}p{1.3em}p{1.3em}p{1.3em}p{1.3em}p{1.3em}p{1.3em}}
    \toprule
          & \textbf{$\alpha$} & \textbf{R1}    & \textbf{R2}    & \textbf{RL}    & \textbf{B.S.}  & \multicolumn{1}{p{4.19em}}{\textbf{Cons. Prop.}} \\
    \midrule
          & base  & 42.0  & 19.3  & 38.6  & 63.5  & 46.0  \\
    \multirow{7}[1]{*}{CNNDM} & 0.05  & 42.1  & 19.2  & 38.7  & 63.5  & 56.6  \\
          & 0.1   & 42.0  & 19.1  & 38.6  & 63.4  & 63.3  \\
          & 0.2   & 41.7  & 18.8  & 38.4  & 63.2  & 70.5  \\
          & 0.5   & 41.6  & 18.7  & 38.3  & 63.1  & 71.8  \\
          & 1     & 41.6  & 18.7  & 38.3  & 63.1  & 71.8  \\
          & 2     & 41.6  & 18.7  & 38.3  & 63.1  & 71.8  \\
          & 5     & 41.6  & 18.7  & 38.3  & 63.1  & 71.8  \\
    \midrule
          & base  & 45.0  & 25.0  & 40.6  & 63.4  & 57.8  \\
    \multirow{7}[1]{*}{NYT} & 0.05  & 45.1  & 25.1  & 40.7  & 63.6  & 58.7  \\
          & 0.1   & 45.1  & 25.1  & 40.7  & 63.5  & 60.2  \\
          & 0.2   & 45.0  & 24.9  & 40.6  & 63.4  & 63.1  \\
          & 0.5   & 44.8  & 24.8  & 40.4  & 63.4  & 65.3  \\
          & 1     & 44.8  & 24.8  & 40.4  & 63.4  & 65.3  \\
          & 2     & 44.8  & 24.8  & 40.4  & 63.4  & 65.3  \\
          & 5     & 44.8  & 24.8  & 40.4  & 63.4  & 65.3  \\
    \midrule
          & base  & 34.5  & 18.5  & 29.1  & 71.9  & 61.6  \\
    \multirow{7}[1]{*}{CNewSum} & 0.05  & 34.5  & 18.5  & 29.1  & 71.9  & 61.7  \\
          & 0.1   & 34.4  & 18.4  & 29.0  & 71.8  & 66.2  \\
          & 0.2   & 34.4  & 18.4  & 29.0  & 71.8  & 66.2  \\
          & 0.5   & 34.4  & 18.4  & 29.0  & 71.8  & 66.2  \\
          & 1     & 34.4  & 18.4  & 29.0  & 71.8  & 66.2  \\
          & 2     & 34.4  & 18.4  & 29.0  & 71.8  & 66.2  \\
          & 5     & 34.4  & 18.4  & 29.0  & 71.8  & 66.2  \\
    \bottomrule
    \end{tabular}%
    \caption{Evaluation performance of checkpoints selected using the different combination ratio $\alpha$ of consecutive proportions to form \textit{S-score}.}
  \label{tab:A1}}%
\end{table}%

\section{Effect of the ratios of $L_{dis}$ and $L_{cohe}$} 
Next, we perform an experiment with varying the combination ratios of $L_{dis}$ and $L_{cohe}$ on CNNDM and CNewSum, which are shown in Table~\ref{tab:A3}. It can found that in most cases, our multitask models provide an improvement in terms of \textit{Cons. Prop.} and \textit{S-score}. The best performance is achieved in the case of $\lambda_{dis} = 0.2$ and $\lambda_{cohe} = 1.0$ on CNNDM, and in the case of $\lambda_{dis} = 0.4$ and $\lambda_{cohe} = 1.0$ on CNewSum, which actually are not our default settings in the main reported experiments ($\lambda_{dis} = 1.0$ and $\lambda_{cohe} = 1.0$). Overall, this ablation study demonstrates that our approach is robust to the hyper-parameter setting of $\lambda$.
We believe the ability of such coherence enhancement is not relying on searching certain values of hyper-parameters, but can be achieved with a large range of hyper-parameters.
\begin{table*}[htbp]
  \centering{\footnotesize
    \begin{tabular}{p{3em}|p{1.3em}p{1.3em}p{1.3em}p{1.3em}p{1.3em}p{1.3em}|p{1.3em}p{1.3em}p{1.3em}p{1.3em}p{1.3em}p{1.3em}} 
    \toprule
    \multirow{2}[4]{*}{Ratio} & \multicolumn{6}{c|}{\textbf{CNewSum}}                  & \multicolumn{6}{c}{\textbf{CNNDM}} \\
\cmidrule{2-13}          & \textbf{R1}    & \textbf{R2}    & \textbf{RL}    & \textbf{B.S.}  & \textbf{Prop. Conti.} & \multicolumn{1}{c|}{\textbf{S-score}} & \textbf{R1}    & \textbf{R2}    & \textbf{RL}    & \textbf{B.S.}  & \textbf{Prop. Conti.} & \multicolumn{1}{c}{\textbf{S-score}} \\
    \midrule
    base  & 34.5  & 18.5  & 29.1  & 71.9  & 61.6  & 215.5  & 42.0  & 19.3  & 38.6  & 63.5  & 46.0  & 209.4  \\
    0.2-1.0 & 34.5  & 18.5  & 29.1  & 71.8  & 61.0  & 214.9  & 41.4  & 18.5  & 38.0  & 62.9  & 75.4  & \textbf{236.2 } \\
    0.4-1.0 & 34.2  & 18.3  & 28.8  & 71.6  & 68.9  & \textbf{221.8 } & 41.5  & 18.6  & 38.2  & 63.0  & 73.5  & 234.9  \\
    0.6-1.0 & 34.5  & 18.5  & 29.1  & 71.8  & 60.6  & 214.5  & 41.5  & 18.7  & 38.2  & 63.1  & 73.1  & 234.6  \\
    0.8-1.0 & 34.5  & 18.5  & 29.1  & 71.9  & 60.9  & 214.9  & 41.6  & 18.7  & 38.2  & 63.1  & 72.5  & 234.1  \\
    1.0-0.2 & 34.6  & 18.6  & 29.2  & 71.8  & 61.7  & 215.9  & 41.8  & 19.0  & 38.5  & 63.3  & 66.1  & 228.7  \\
    1.0-0.4 & 34.6  & 18.5  & 29.1  & 71.8  & 63.2  & 217.2  & 41.7  & 18.9  & 38.4  & 63.2  & 69.2  & 231.4  \\
    1.0-0.6 & 34.6  & 18.5  & 29.1  & 71.8  & 63.4  & 217.4  & 41.7  & 18.8  & 38.3  & 63.1  & 71.2  & 233.0  \\
    1.0-0.8 & 34.6  & 18.5  & 29.1  & 71.9  & 62.1  & 216.1  & 41.6  & 18.7  & 38.3  & 63.1  & 71.5  & 233.2  \\
    1.0-1.0 & 34.4  & 18.4  & 29.0  & 71.8  & 66.2  & 219.9  & 41.5  & 18.4  & 38.1  & 62.9  & 72.8  & 233.6  \\
    \bottomrule
    \end{tabular}%
    \caption{Comparison with different ratios of two loss function on CNNDM and CNewSum datasets. }
  \label{tab:A3}}%
\end{table*}%

\section{Effect of the ratios of $L_{ext}$} 

Another experiment is conducted to check the effect of $L_{ext}$ in multitask fine-tuning. We fix $\lambda_{dis}=\lambda_{cohe}=1.0$ and change $\lambda_{ext}$ from 0 to 1 with a stepsize of 0.2.
The results are given in Table ~\ref{tab:A4}. It shows that increasing $\lambda_{ext}$ from 0.2 to 1.0 is not helpful in improving the consecutive proportion and S-score, although the values of relevance metrics are preserved. This may due to the gradient conflict effect of three losses.
\begin{table*}[htbp]
  \centering{\footnotesize
    \begin{tabular}{p{3.0em}|p{1.3em}p{1.3em}p{1.3em}p{1.3em}p{1.3em}p{2.0em}|p{1.3em}p{1.3em}p{1.3em}p{1.3em}p{1.3em}p{2.0em}}
    \toprule
    \multirow{2}[4]{*}{\textbf{Ratio}} & \multicolumn{6}{c|}{\textbf{CNNDM}}                    & \multicolumn{6}{c}{\textbf{CNewSum}} \\
\cmidrule{2-13}          & \textbf{R1}    & \textbf{R2}    & \textbf{RL}    & \textbf{B.S.}  & \textbf{Prop. Conti.} & \textbf{S-score} & \textbf{R1}    & \textbf{R2}    & \textbf{RL}    & \textbf{B.S.}  & \textbf{Prop. Conti.} & \textbf{S-score} \\
    \midrule
    base   & 42.0  & 19.3  & 38.6  & 63.5  & 46.0  & 209.4 & 34.5  & 18.5  & 29.1  & 71.9  & 61.6  & 215.5  \\
    0     & 41.5  & 18.4  & 38.1  & 62.9  & 72.8  & \textbf{233.6}  & 34.4  & 18.4  & 29.0  & 71.8  & 66.2  & \textbf{219.9 } \\
    0.2   & 42.0  & 19.2  & 38.5  & 63.4  & 43.5  & 206.6  & 34.4  & 18.4  & 28.9  & 71.8  & 61.6  & 215.0  \\
    0.4   & 42.0  & 19.2  & 38.5  & 63.4  & 43.5  & 206.6  & 34.4  & 18.4  & 28.9  & 71.8  & 62.1  & 215.5  \\
    0.6   & 42.0  & 19.2  & 38.5  & 63.4  & 43.5  & 206.6  & 34.4  & 18.4  & 28.9  & 71.8  & 62.1  & 215.6  \\
    0.8   & 42.0  & 19.2  & 38.5  & 63.4  & 43.5  & 206.6  & 34.4  & 18.4  & 28.9  & 71.8  & 62.1  & 215.6  \\
    1     & 42.0  & 19.2  & 38.5  & 63.4  & 43.5  & 206.7  & 34.4  & 18.4  & 28.9  & 71.9  & 61.7  & 215.3  \\
    \bottomrule
    \end{tabular}%
    \caption{Comparison with different ratios of $L_{ext}$ in multitask finetuning.}
  \label{tab:A4}}%
\end{table*}%

\section{Graphical Analysis} \label{Sec:graph_analysis}

\begin{figure*}[h]
\begin{center}
 \includegraphics[width=1.0\linewidth]{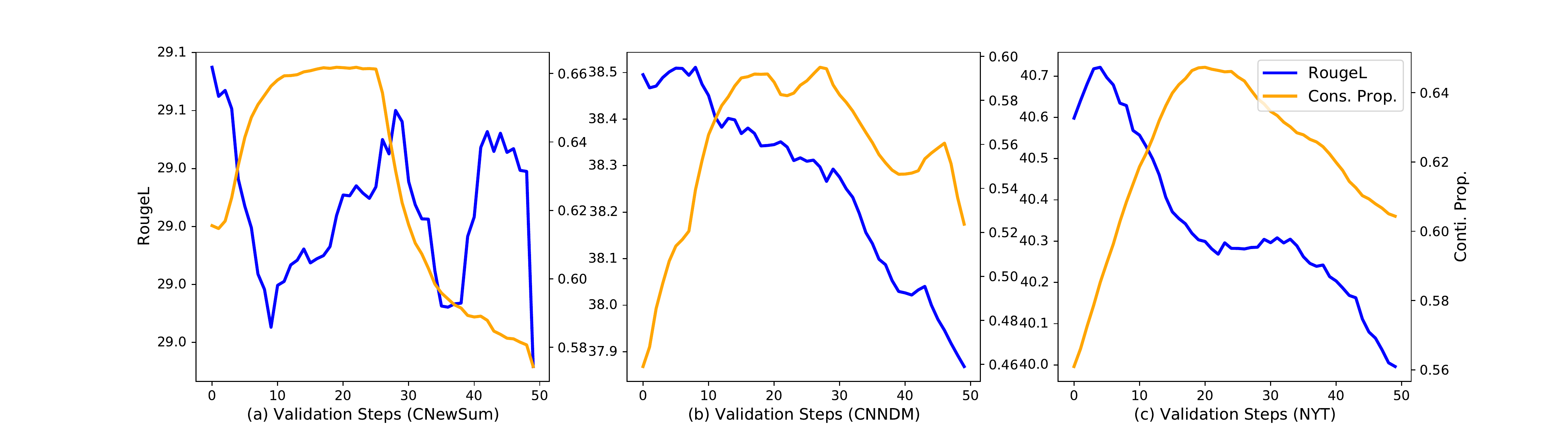}
 \caption{Learning curves of Rouge-L and Consecutive Proportion of three datasets. 
 For CNewSum (a) we apply \textit{model-based} setting, while for CNNDM (b) and NYT (c) we apply \textit{MAT-based} setting, as was discussed in Section~\ref{Sec:4.4.1}.
 } \label{Fig:cohe_finetune}
\vspace{-1.5em}
\end{center}
\end{figure*}
\begin{figure*}[h]
\begin{center}
 \includegraphics[width=1.0\linewidth]{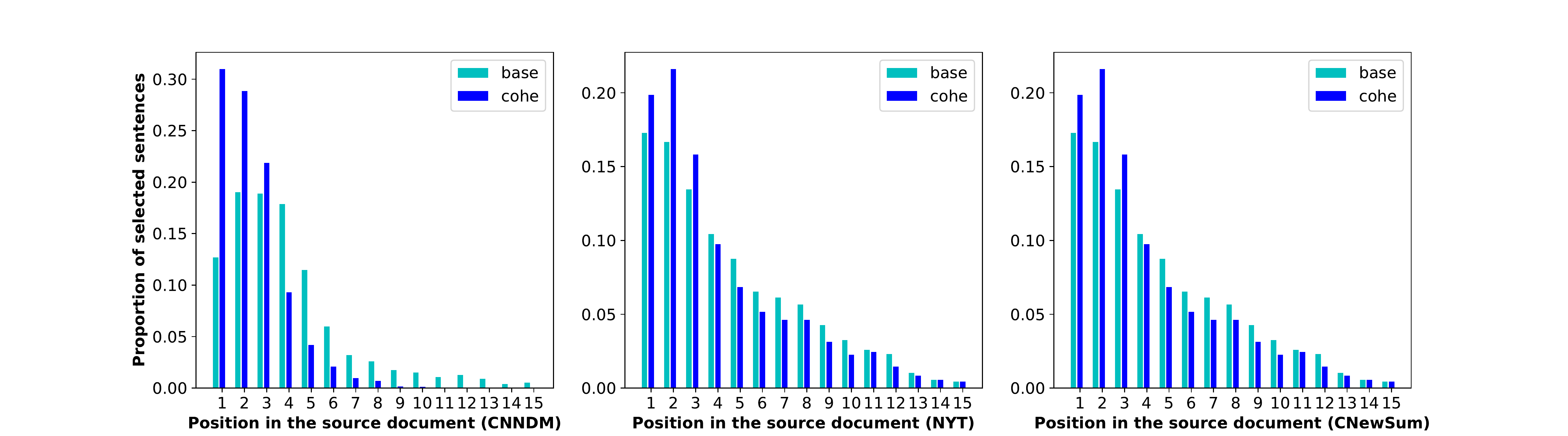}
 \caption{Comparison of summary sentence positions in the source document. Summary sentences are extracted separately by the baseline model (denoted by \texttt{base}) and our model with coherence enhancement (denoted by \texttt{cohe}). } \label{Fig:sents_pos}
\end{center}
\end{figure*}

\noindent \textbf{Learning curves}. We further investigate the learning curves of the losses as well as the evaluation metrics. As is shown in Figure~\ref{Fig:cohe_finetune}, in general, the value of RougeL decreases slowly as the fine-tuning process moves on. However, the proportion of consecutive sentences (\textit{Cons. Prop.}) increases sharply at the beginning steps, which contributes to enhance the coherence of result summaries. \\
\noindent \textbf{Positional Distributions}. We also make a comparison of the position distribution of summary sentences extracted by the model before and after multitask fine-tuning on the three datasets, as given in Figure \ref{Fig:sents_pos}. It is shown that for CNNDM, the position distribution of extracted sentences after coherence enhancement has a significant shift from the baseline distribution, and gathers more at the front of the document. This may result in a higher consecutive proportion and coherence, but sacrifice rouge scores when compared to the reference summary. For NYT and CNewSum, the shift is much smaller, while correspondingly, the consecutive proportion shown in Table~\ref{tab:merge} are not increased as much as the case of CNNDM.

\end{document}